\documentclass[]{spie}  

\usepackage{amsmath,amsfonts,amssymb}
\usepackage{graphicx}
\usepackage[colorlinks=true, allcolors=blue]{hyperref}
\usepackage{multiraw}
\usepackage[english, russian]{babel}  

\title{Handwritten text generation and strikethrough characters augmentation}

\author[a]{Alex Shonenkov} 
\author[b]{Denis Karachev}
\author[c]{Max Novopoltsev}

\author[d]{Mark Potanin}
\author[e]{Denis Dimitrov}
\author[f]{Andrey Chertok}
\affil[a,c]{SBER AI, Moscow, Russia}
\affil[b]{OCRV, Sochi, Russia}
\affil[d]{SBER AI, MIPT, Moscow, Russia}
\affil[e]{SBER AI, Lomonosov MSU, Moscow, Russia}
\affil[f]{SBER AI, AIRI, Moscow, Russia}
\authorinfo{Further author information: \\Alex Shonenkov.: E-mail: shonenkov@phystech.edu\\Denis Karachev.: E-mail:Denis.Karachev@OCRV.ru\\ Max Novopoltsev.: E-mail: MYNovopoltsev@sberbank.ru\\ Mark Potanin.: E-mail: mark.potanin@phystech.edu\\Denis Dimitrov.: E-mail: denis.dimitrov@math.msu.su\\Andrey Chertok.: E-mail:AChertok@sberbank.ru}

\begin{document} 
\maketitle
\selectlanguage{english}
\begin{abstract}
We introduce two data augmentation techniques, which, used with a Resnet - BiLSTM - CTC network, significantly reduce Word Error Rate (WER) and Character Error Rate (CER) beyond best-reported results on handwriting text recognition (HTR) tasks. We apply a novel augmentation that simulates strikethrough text (HandWritten Blots) and a handwritten text generation method based on printed text (StackMix), which proved to be very effective in HTR tasks. StackMix uses weakly-supervised framework to get character boundaries. Because these data augmentation techniques are independent of the network used, they could also be applied to enhance the performance of other networks and approaches to HTR. Extensive experiments on ten handwritten text datasets show that HandWritten Blots augmentation and StackMix significantly improve the quality of HTR models.
\end{abstract}

\section{INTRODUCTION}
\label{sec:intro}  

Handwriting text recognition (HTR) is a vital task. Automation allows for a dramatic reduction in labor costs for processing correspondence and application forms and deciphering historical manuscripts. The main problem with historical documents is the usual small amount of labeled data. However, HTR systems require many examples for training and setting parameters. Optical character recognition (OCR) is an area in which deep learning has proven itself perfectly. The situation in HTR, especially with historical documents, which are much worse. Since there are only a few open datasets, the quality of trained models recognition is much lower.

To improve the state-of-the-art of HTR system, we introduce two ways to increase the volume of training data: augmentation that simulates strikethrough text - HandWritten Blots and a "new text" generation method - StackMix.

\begin{figure} [ht]
   \begin{center}
   \begin{tabular}{c} 
   \includegraphics[width=0.8\linewidth]{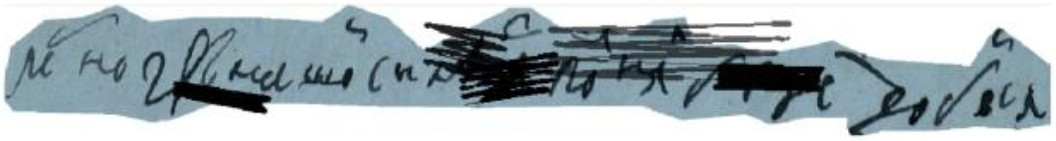}
   \end{tabular}
   \end{center}
   \caption 
   {\label{fig:hwb-hwb-example}Sample handwritten text image after using HandWritten Blots.}
   \end{figure}

\begin{figure} [ht]
   \begin{center} 
   \begin{tabular}{c} 
   \includegraphics[width=0.7\linewidth]{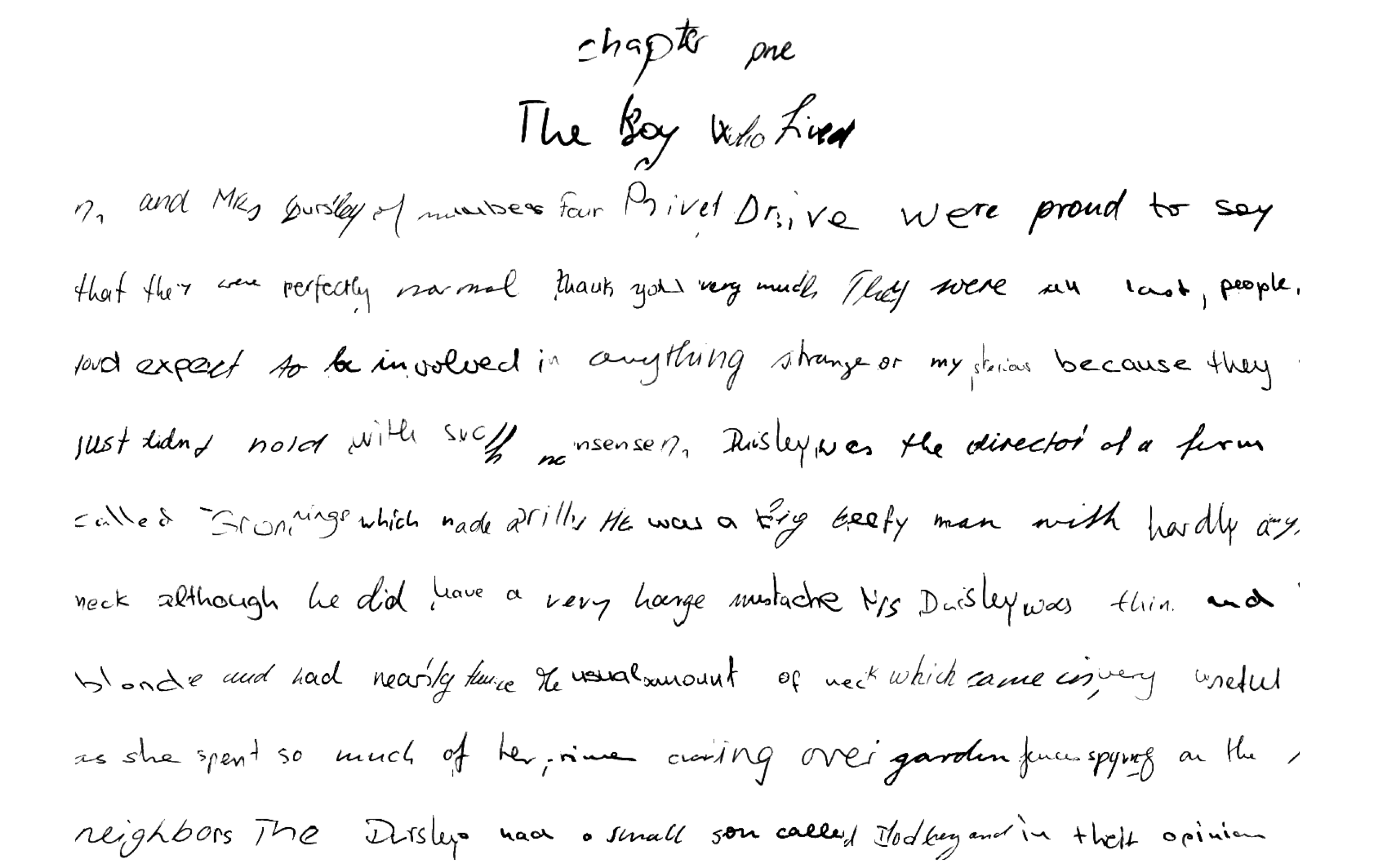}
   \end{tabular} 
   \end{center}
   \caption {\label{fig:hp-iam}Example first page of Harry Potter created using  StackMix from IAM style.}
   \end{figure}

\begin{figure} [ht]
   \begin{center} 
   \begin{tabular}{c} 
   \includegraphics[width=0.7\linewidth]{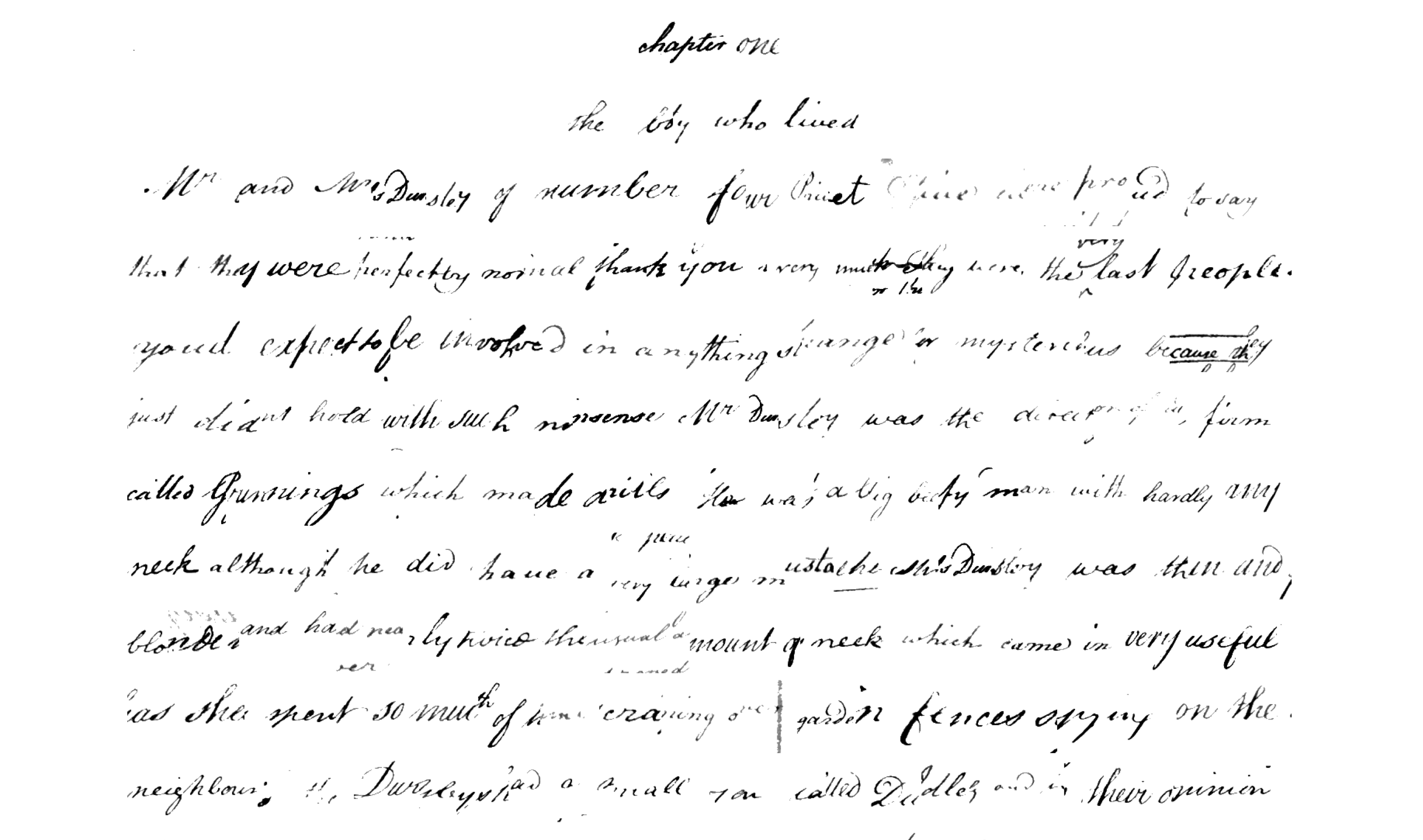}
   \end{tabular} 
   \end{center}
   \caption {\label{fig:hp-bentham}Example first page of Harry Potter created using  StackMix from Bentham style}
   \end{figure}

The proposed HandWritten Blots simulates the strikethrough characters as close as possible to the originals. It can change the inclination, size, and transparency of drawing lines that strike out characters. This is illustrated in Fig.\ref{fig:hwb-hwb-example}. Stackmix generates handwritten text (phrase, string, or entire page) using characters images of the training dataset. We proposed weakly-supervised learning to extract characters boundaries from training images. As an example, we generated pages of texts from different sources. Some paragraphs from the first chapter of a Harry Potter book were used for all models (Fig.\ref{fig:hp-iam}, \ref{fig:hp-bentham}). The results suggest that it is possible to generate different texts with different styles and fonts. 


These augmentations were designed initially for the system to decipher Peter the Great manuscripts that were first introduced at
\cite{potanin2021digital} by using marked-up lines of the text as input.

We evaluate our augmentation and normalization techniques
using a Resnet - BiLSTM - CTC architecture to perform HTR. Our approach achieves low  Word Error Rate (WER) and Character Error Rate(CER) on ten different handwritten texts dataset. This dataset consists of handwritten texts over thousands of authors and multiple languages written centuries apart.




\section{RELATED WORK}
\subsection{Data Augmentation}
Data augmentation consists of augmenting the training set with synthetically generated samples. It reduces the tendency to overfit when training models with many parameters and limited labeled data. In data augmentation for image classification problems, the training set is increased by modifying the original images through transformations such as scaling, rotation, or flipping images and generating new images from part of the images of the training dataset.

For example, in the CutMix approach \cite{yun2019cutmix}, parts of the images are cut from different samples and inserted into a new one. At the same time, the targets are mixed according to the proportions of the original parts of the images. A similar approach is used in SnapMix \cite{huang2020snapmix}, but the cutting of images is regulated using Class Activation Map. This approach allows for reducing the noise of the cut objects and selecting the most significant parts. Additionally, an exciting approach with mixing objects is presented in MixUp\cite{zhang2018mixup}, and MWH \cite{yu2021mixup}, where images overlap each other with a transparency coefficient and mix the targets with the proportion. Unfortunately, these methods cannot be applied to the optical character recognition and handwritten text recognition tasks because mixing recursively dependent targets makes it very difficult to obtain a correct mapping of image and text. Therefore, we would like to introduce the StackMix approach, which improves the quality and stability of our neural network.

Several authors have proposed specific augmentation techniques for HTR. In  \cite{wigington2017data}  the authors introduce a more robust augmentation technique and normalization to model the handwritten text variation of a given author. In\cite{Poznanski2016}, they show some affine transformation methods for data augmentation in HTR. \cite{krishnan2016matching} and \cite{shen2016method} synthesize new lines images by concatenating characters from different datasets. In \cite{krishnan2016matching}, the authors introduce a new method for matching double-sided historical documents to remove interference caused by handwriting from the reverse side due to ink sipping over long storage periods. This article proposes different strategies for obtaining synthetic handwritten Chinese documents using an existing segmented database at the character level. In \cite{chammas2018handwriting}, the authors improve the performance by augmenting the training set with specially crafted multiscale data.

Common tricks may significantly improve the quality of HTR models. In \cite{aradillas2020boosting}, the authors investigated data augmentation and transfer learning for small historical datasets. 

Obtaining a massive corpus of labeled handwriting images for different languages is a cumbersome task. Authors use Generative Adversarial Networks (GAN) (called ScrabbleGAN) to generate training data \cite{fogel2020scrabblegan}. ScrabbleGAN follows a semi-supervised approach to synthesize handwritten text images that are versatile both in style and lexicon. It can generate images of varying lengths. The generator can also manipulate the resulting text style, allowing us to decide whether the text must be cursive or how thick/thin the pen stroke should be.

\subsection{Handwritten text Recognition systems}

Handwritten text recognition is a long-standing computer vision problem. Early works on handwritten recognition problems suggest using a combination of Hidden Markov Models (HMM) and recurrent neural networks (RNNs) \cite{bengio1999, bourlard1994}. 
The disadvantage of these approaches is the impossibility of end-to-end loss function optimization. With the advent of deep learning came tremendous improvements in handwriting recognition.

Poznaski and Wolf \cite{Poznanski2016} perform word-level recognition by employing a fixed-size CNN architecture that evaluates binary lexical attributes over word images, such as whether a given portion of the image contains a certain unigram or bigram (PHOC \cite{almazan2014word}). The correct transcription is determined by the word in the lexicon closest to this representation. Authors \cite{krishnan2016deep} also employ a fixed-size CNN architecture to learn features for the PHOC representation for embedding the text and images into a common subspace.

Another general approach uses RNNs for HTR. These have been widely adopted with the introduction of Connectionist Temporal Classification (CTC). CTC is an algorithm used to deal with tasks like speech recognition, handwriting recognition, where just the input data and the output transcription are available, but there are no characters boundaries provided.
The basic idea was to interpret the network outputs as a probability distribution over all possible label sequences conditioned on a given input sequence. Given this distribution, an objective function can be derived that directly maximizes the probabilities of the correct labelings. 

State-of-the-art HTR architectures combine a convolutional neural network (CNN)
with Long Short-Term Memory Recurrent Neural Networks (LSTM) cells \cite{hochreiter1997long}.
This type of network model can deal with sequential data to identify temporal patterns.

Large Multidimensional Long Short-Term Memory Recurrent Neural Networks (MDLSTM) \cite{voigtlaender2016handwriting} networks use 2D-RNN, which can deal with both axes of an input image. A simple model consists of several convolutional neural network (CNN) and MDLSTM layers, and using connectionist temporal classification (CTC) loss provides excellent metrics for the IAM \cite{marti2002iam} dataset.

However, MDLSTM models have some disadvantages like high computational costs and instability. In works  \cite{coquenet2020recurrence} and \cite{ingle2019scalable}, the authors try to eliminate recurrent layers in CNN-LSTM-CTC to decrease the number of parameters. Their Gated Fully Convolutional Networks show relatively good results, even without a language model.Another alternative to the RCNN-CTC approach is seq2seq models \cite{michael2019evaluating}. The encoder extracts features from the input, and the decoder with an attention mechanism emits the output sequentially. OrigamiNet \cite{yousef2020origaminet} is a module that combines segmentation and text recognition tasks. It can be added to any line-level handwritten text recognition model to make it page-level.

We compared the proposed model with the models described above. The data for comparison is given in section \ref{sec:results}. We also found that different authors used various data partitions for train, test, and validation on the IAM dataset (see section \ref{sec:dataset}). This made it difficult to compare the models correctly, so we present our model results for the same data sets used in the original papers.


\section{PROPOSED AUGMENTATIONS ALGORITHMS}
\subsection{Handwritten Blot Augmentation}

The idea of augmentation appeared during the analysis of the Digital Peter dataset \cite{complink,potanin2021digital}. In the process of examining the dataset, we found examples of images in which some characters were crossed out and almost indistinguishable, but they were still present in the markup. Hence, the idea of using the Cutout augmentation \cite{devries2017improved} emerged since it allows for overlapping of some elements of symbols or entire symbols, which makes the augmentation a bit like crossed-out symbols. 

However, in training the models, we decided to implement such an algorithm that would allow for simulating the strikethrough characters as close as possible to the originals. Since we did not find the implementation of such algorithms in open sources, we created it ourselves. We significantly improve the quality of HTR models using Handwritten Blot augmentation than using Cutout augmentation. 

To implement the strikethrough effect, we decided to use the Bezier curve construction algorithm, which in our case smoothed the curve transition between points. The Bezier curve is a parametric curve and is a special case of the Bernstein polynomial. Finding basic polynomials of degree n are found by the formula:

\begin{equation}
\small\label{eq4}
{b_{j,n}}=\left(\begin{array}{c}n\\j\end{array}\right)s^j\left(1-s\right)^{n-j}
\end{equation}

 Where $j=0,\dots ,n$. The definition of a curve as a linear combination is found as:

\begin{equation}
\small\label{eq5}
B(s)=\sum_{j=0}^nb_{j,n}\cdot{v_j}
\end{equation}

Where $v_{j}$ is the point in the space, and $b_{j, n}$ define above. Since the sum of all polynomials must be equal to one, then.

\begin{equation}
\small\label{eq6}
b_{0,n}+b_{0,1}+...+b_{n,n}=(s+(1-s))^n=1
\end{equation}

Where S is non-negative weights that sum to one. We found the implementation of the algorithm for constructing the Bezier curve in \cite{Hermes2017}.

\begin{figure} [ht]
   \begin{center}
   \begin{tabular}{c} 
    \includegraphics[width=0.5\linewidth]{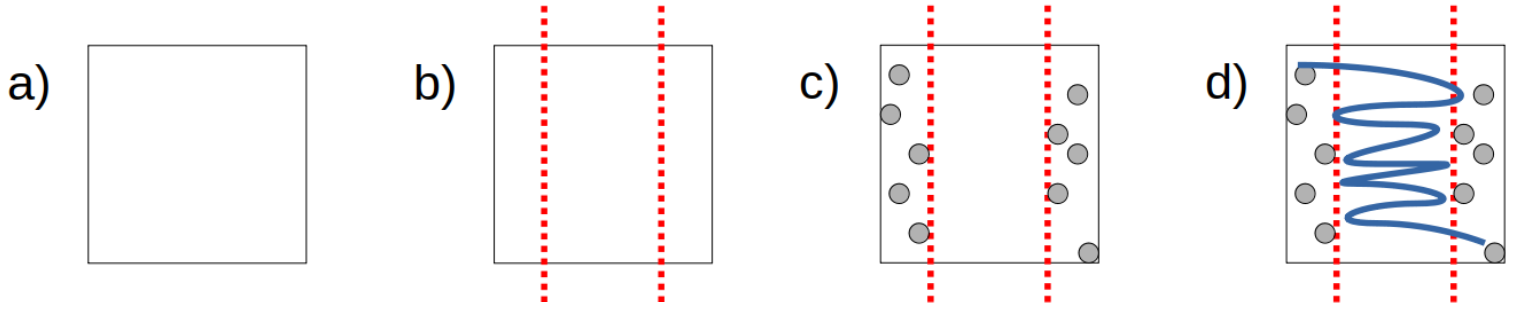}
   \end{tabular}
   \end{center}
   \caption {\label{fig:hwb-algorithm-explanation}Graphic description of the algorithm.}
   \end{figure}

Next, we implemented our algorithm that simulates strikethrough. A graphical description of the algorithm is shown in Fig. \ref{fig:hwb-algorithm-explanation}. The main steps were as follows:
\begin{itemize}
    \item Determine the coordinates of the strikethrough area (Fig.\ref{fig:hwb-algorithm-explanation}, a).
    \item Define areas for generating points to be used for drawing a Bezier curve (Fig.\ref{fig:hwb-algorithm-explanation}, b).
    \item Generate points for the Bezier curve with the parameters of the intensity of the points and their coordinates to simulate the slope. Sometimes, a random point needs to be used several times for the loop to go slightly further from the curve (Fig.\ref{fig:hwb-algorithm-explanation}, c). 
    \item Draw a curve with specified transparency (Fig.\ref{fig:hwb-algorithm-explanation}, d). 
\end{itemize}

The implementation of this algorithm can be found here \cite{codelink}. We empirically selected the parameters for strikethrough (minh = 50, maxh = 100, minw = 10, maxw = 50, incline = 15, intensity = 0.9, transparency = 0.95, count = 1 \ldots 11, proba = 0.5) and tested it on different datasets.These parameters can change the inclination, transparency, and size of drawing lines that strikeout characters.

The effect of the Blot augmentation on quality metrics is shown in the "blots" row of Table \ref{tab:extended-all-results-table} and in Figure \ref{fig:test_cer}.

The obtained data suggest that handwritten blot augmentation makes a significant contribution to the quality of training models. Therefore, we recommend using it to train models in handwriting recognition problems.

\subsection{Handwriten text generation by StackMix algorithm}
\begin{figure} [ht]
   \begin{center}
   \begin{tabular}{c} 
    \includegraphics[width=0.7\linewidth]{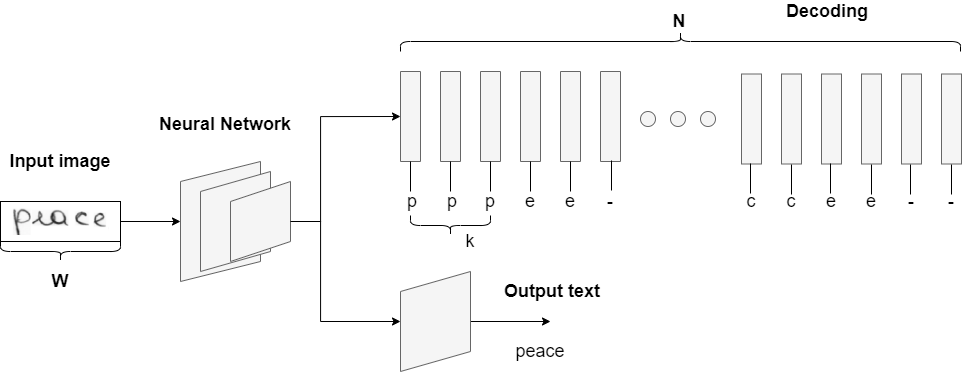}
   \end{tabular}
   \end{center}
   \caption {\label{fig:char-masks}Post-processing scheme to get the boundaries of the symbols.}
   \end{figure}

Proposed Handwritten text generation algorithm StackMix generates synthetic lines of text, given samples of the isolated characters. We used weakly-supervised learning to extract characters boundaries from training images. The algorithm is based on the post-processing of a supervised pretrained neural network via CTC loss. It gets characters boundaries using only weakly-supervised learning without any manual markup (Fig. \ref{fig:char-masks}).  The training does not require character-level annotation and can be applied to many cases where only labeled words or entire phrases samples are available. During training the order of characters is given, and the goal is to separate the input image by these characters, proportional to the size of the output sequence. We use the intersection coordinates of the different characters to get the coordinates of the beginning and end of each character. We can generate any phrase from different datasets style with all characters coordinates in all the lines.

\begin{figure} [ht]
   \begin{center}
   \begin{tabular}{c} 
    \includegraphics[width=0.2\linewidth]{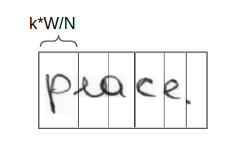}
   \end{tabular}
   \end{center}
   \caption {\label{fig:char-masks_2}The boundaries of the symbols.}
   \end{figure}
\begin{figure} [ht]
   \begin{center}
   \begin{tabular}{c} 
    \includegraphics[width=0.9\linewidth]{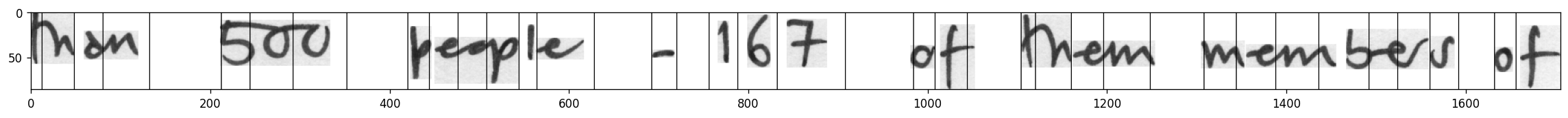}
   \end{tabular}
   \end{center}
   \caption {\label{fig:example1}Example images of symbol segmentation using semi-supervised methods. Images have ids: IAM-a02-082-05.}
   \end{figure}

Character width is proportional to k*W / N, where k - number of "cells" with maximum character probability; W - input images width; N - number of samples (Fig. \ref{fig:char-masks_2}). The main idea was to connect the last layer of RNN (after applying SoftMax activation for every character from image features) and image width. For training neural network can be used base scheme without any augmentations and tricks. To get high-quality marking of symbols boundaries, a sample from the training stage should be used. Example of image of symbol segmentation using weakly-supervised method for IAM dataset are given in Fig. \ref{fig:example1}).

The input for the algorithm is expected to be text from the external corpus, which creates a new image with this text using parts of images of the training dataset. The algorithm from natural language toolkit for python (nltk) \cite{bird2009natural} MWETokenizer \cite{nltkmwe} is used for tokenization. It processes tokenized text and merges multi-word expressions into single tokens. Collections of multi-word expressions are obtained from the training dataset, using symbol borders to connect parts of images and MWE tokens, including spaces and punctuation marks. In this study, we used one random MWE tokenizer from six with a token dimension of no more than 3, 4, 5, 6, 7, and 8 with probabilities 0.05, 0.15, 0.2, 0.2, 0.2, and 0.2, respectively. After this, each token was matched with part of an image from the training data. The pieces were stacked (hence the name "StackMix") together into a complete image, maintaining the correct order of the tokens.   

The StackMix approach also requires an external text corpus that has the same alphabet as the main dataset. Corpus does not require special marking and only contains allowed symbols. We use different text corpus over thousands of authors and multiple languages written
centuries apart corresponding to each dataset in Table \ref{tab:text_corpus}.

\begin{table}[ht]
\caption{Text corpus for different datasets.} 
\label{tab:text_corpus}
\begin{center} 
    \begin{tabular}{ |l|c|c| }
    \hline
    \textbf{Dataset} &  \textbf{Text corpus}  \\
    \hline
    Bentham, IAM & "Jigsaw Unintended Bias in Toxicity Classification" \cite{jigsaw} \\
    \hline
    Saint Gall & "The Latin Library" \cite{the_latin_library}\\
    \hline
    HKR & Russian texts from Wikimedia~\cite{ruwiki}\\
    \hline
    Digital Peter & russian texts of the XVII-XVIII centuries \\
    \hline
    \end{tabular}
    \end{center}
\end{table}

\begin{figure} [ht]
   \begin{center}
   \begin{tabular}{c} 
    \includegraphics[width=0.7\linewidth]{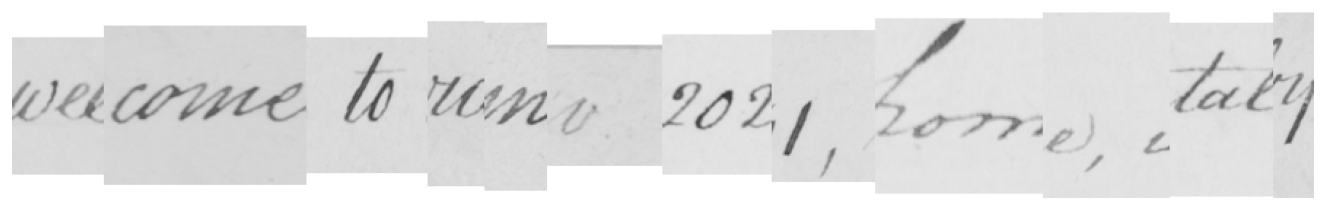}
   \end{tabular}
   \end{center}
   \caption {\label{fig:example2_c}Stackmix example line from the Bentham.\textit{""Welcome to ICMV 2021, Rome, Italy."}}
   \end{figure}

\begin{figure} [ht]
   \begin{center}
   \begin{tabular}{c} 
    \includegraphics[width=0.7\linewidth]{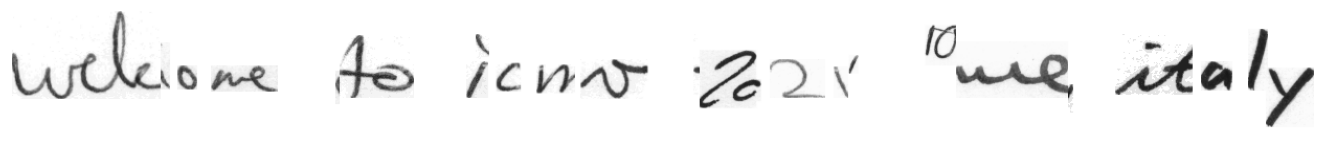}
   \end{tabular}
   \end{center}
   \caption {\label{fig:example1_c}Stackmix example line from the IAM.\textit{"Welcome to ICMV 2021, Rome, Italy."}}
   \end{figure}

\begin{figure} [ht]
   \begin{center}
   \begin{tabular}{c} 
    \includegraphics[width=0.7\linewidth]{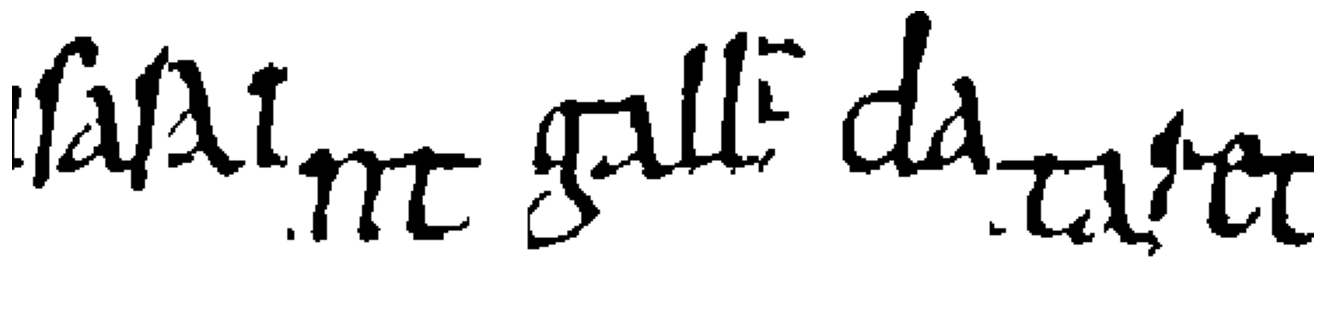}
   \end{tabular}
   \end{center}
   \caption {\label{fig:example3_c}Stackmix example line from the Saint Gall. \textit{"SaintGall dataset."}}
   \end{figure}

\begin{figure} [ht]
   \begin{center}
   \begin{tabular}{c} 
    \includegraphics[width=0.7\linewidth]{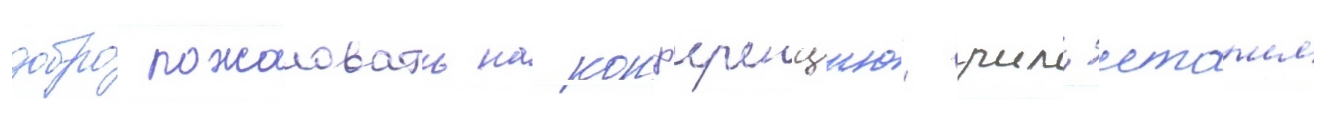}
   \end{tabular}
   \end{center}
   \caption {\label{fig:example4_c}Stackmix example line from the HKR.\selectlanguage{russian}\textit{"Добро пожаловать на конференцию, Рим, Италия."}}
   \end{figure}

\begin{figure} [ht]
   \begin{center}
   \begin{tabular}{c} 
    \includegraphics[width=0.7\linewidth]{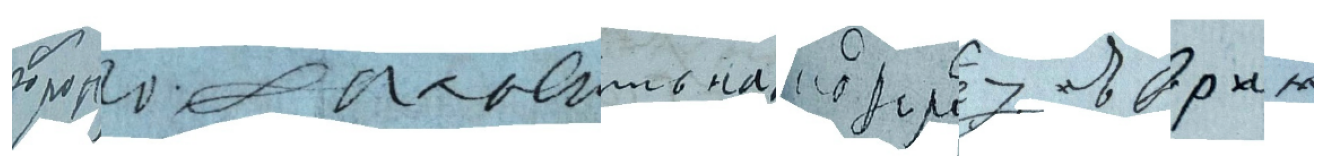}
   \end{tabular}
   \end{center}
   \caption{\label{fig:example5_c}Stackmix example line from the Peter.\selectlanguage{russian}\textit{"Добро пожаловать на конференцию в Рим."}} 

\end{figure}

Examples of the StackMix algorithm for various datasets are given in Figures \ref{fig:example2_c}--\ref{fig:example5_c}. Experiments result show that StackMix augmentation improves the quality of training models (Table \ref{tab:extended-all-results-table}, row StackMix). Despite the visible places where tokens were glued together, the algorithm significantly increased the quality of recognition. We tried to increase the realism of the generated text by alignment and selection of samples. However, it did not improve the quality of our experiments.

Nevertheless, after specific improvements, this algorithm may be used for the realistic generation of new documents. 

\section{NEURAL NETWORK ARCHITECTURE AND HANDWRITTEN TEXTS DATASETS}

\subsection{Neural Network Architecture}
The neural network underlying the proposed system consists of three parts: a feature generator, RNN to account for the order of the features, and the classifier that outputs the probability of each character (use CTC Loss).

\begin{figure} [ht]
   \begin{center}
   \begin{tabular}{c} 
   \includegraphics[width=0.8\linewidth]{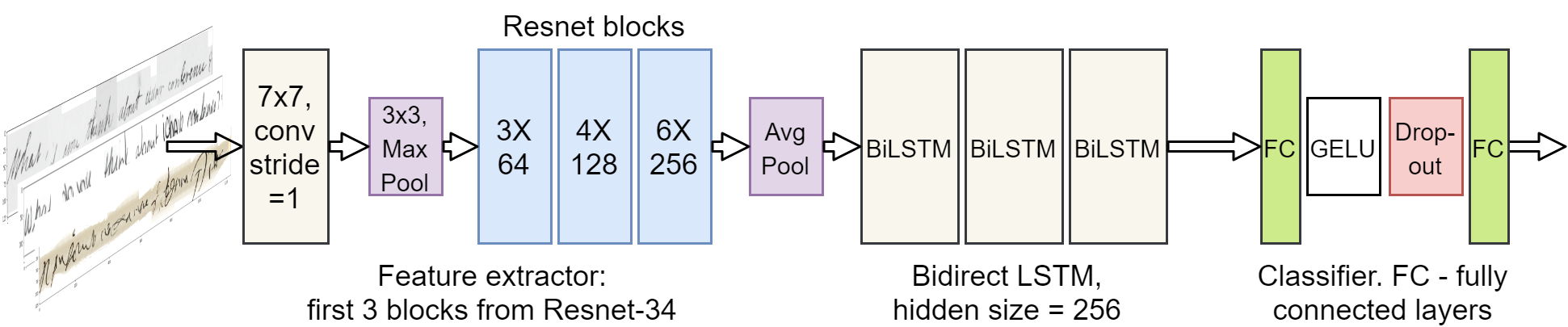}
   \end{tabular}
   \end{center}
   \caption 
   {\label{fig:fig:nn}Neural network architecture.}
   \end{figure}

As a feature generator, various network architectures were tested, and the final choice fell on Resnet (Fig.~\ref{fig:fig:nn}). We took only three first blocks from Resnet-34 and replaced the stride parameter in the first layer with 1 to increase the "width" of the objects.

After the features were extracted, they were averaged through the AdaptiveAvgPool2d layer and fed into the three BiLSTM layers to deal with feature sequences.

As a final classifier, we use two fully connected layers with GELU and dropout between them.

The results achieved using the described architecture without any additional modifications are shown in the "base" row of Table \ref{tab:extended-all-results-table}.

\subsection{Datasets}
\label{sec:dataset}
Ten different datasets were used in the experiments to prove state-of-the-art quality of our model. We have tested our two augmentations techniques over thousands of authors, and multiple languages were written centuries apart.

Bentham manuscripts refers to a large set of documents that were written by the renowned English philosopher and reformer Jeremy Bentham (1748-1832). Volunteers of the Transcribe Bentham\footnote{http://transcribe-bentham.ucl.ac.uk/td/Transcribe\_Bentham} initiative transcribed this collection. 
Currently, $>$ 6 000 documents or $>$ 25 000 pages have been transcribed using this public web platform.

For our experiments, we used the BenthamR0 dataset \cite{bentham} a part of the Bentham manuscripts.

IAM handwriting dataset contains forms of handwritten English text. It consists of 1 539 pages of scanned text from 657 different writers. This dataset is widely used for experiments in many papers. However, there is a big problem with uncertainty in data splitting for model training and evaluation. This issue was described in \cite{michael2019evaluating}. The IAM dataset has different train/val/test splits that are shown in Table \ref{tab:iam_split}. The problem is that none of them are labeled as a standard, so the IAM dataset split differs from paper to paper. However, results should be compared on the same split. 

In our experiments, we use the IAM-B\footnote{http://www.tbluche.com/resources.html} and IAM-D partitions. IAM-B was used to compare our model with others. IAM-D was a new partition inspired by the official page of  project\footnote{https://fki.tic.heia-fr.ch/databases/iam-handwriting-database}. This page contained "unknown","val1" and "val2" split labels. We added "unknown" samples to the train set, and combined "val1" and "val2" together.

IAM-B was chosen because many recently published papers used this partition. We used IAM-D because it provides more training samples.

We create a github repository \cite{iam_splits} with information and indexes corresponding to each IAM split in Table \ref{tab:iam_split}. It also contains links to papers that use these splits. We hope this helps researchers choose appropriate IAM partitions and make valid comparisons with other papers.

\begin{table}[ht]
\caption{IAM splits.} 
\label{tab:iam_split}
\begin{center} 
    \begin{tabular}{ |l|c|c|c| }
    \hline
    \textbf{Split} &  \textbf{Train} & \textbf{Val} & \textbf{Test} \\
    \hline
    IAM-A & 6161 & 966 & 2915 \\
    IAM-B & 6482 & 976 & 2915 \\
    IAM-C & 6161 & 940 & 1861 \\
    IAM-D & 9652 & 1840 & 1861 \\
    \hline
    \end{tabular}
    \end{center}
\end{table}

Saint Gall dataset contains handwritten historical manuscripts written in Latin that date back to the 9th century. It consists of 60 pages, 1 410 text lines and 11 597 words.

HKR \cite{nurseitov2020hkr} is a recently published dataset of modern Russian and Kazakh language. This database consists of $>$ 1 400 filled forms. It contains 64 943 lines and $>$ 715 699 symbols produced by about 200 different writers. Data are split in the following manner: 45 559 lines for a train set, 10 009 lines for validation, and 9 375 lines for test. This data splitting was found in github \cite{hkr_splitting_github} of the authors of the HKR\_Dataset, but these proportions of train/valid/test were slightly different that those of the original paper \cite{nurseitov2020hkr}. We assumed that the seed was not fixed in a script to get split, which was not a big problem for comparing results. 

  \selectlanguage{english}  
Digital Peter is a completely new dataset of Peter the Great's manuscripts \cite{potanin2021digital}. It consists of 9 694 images and text files corresponding to lines in historical documents. The open machine learning competition Digital Peter was held based on the considered dataset \cite{complink}. There are 6 237 lines in the training set, 1 527 lines in the validation set, and 1 930 lines in the testing set. 
  \selectlanguage{english} 
Konzil dataset was created by specialists of the University of Greifswald. It contains manuscripts written in modern German. Train sample consists of 353 lines, validation -- 29 lines and test -- 87 lines.

Schiller contains handwritten texts written in modern German. Train sample consists of 244 lines, validation -- 21 lines and test -- 63 lines.

Ricordi contains handwritten texts written in Italian. Train sample consists of 295 lines, validation -- 19 lines and test -- 69 lines.

Patzig contains handwritten texts written in modern German. Train sample consists of 485 lines, validation -- 38 lines and test -- 118 lines.

Schwerin contains handwritten texts written in medieval German. Train sample consists of 793 lines, validation -- 68 lines and test -- 196 lines.

\section{RESULTS AND DISCUSSION}
\label{sec:results}
The four most frequently used in HTR task datasets and Digital Peter dataset were used in the experiments to prove the state-of-the-art quality of our model. Digital Peter is an entirely new dataset\cite{potanin2021digital}. Authors present detailed information about Digital Peter dataset and share links to data. 

\begin{figure} [ht]
   \begin{center}
   \begin{tabular}{c} 
    \includegraphics[width=0.8\linewidth]{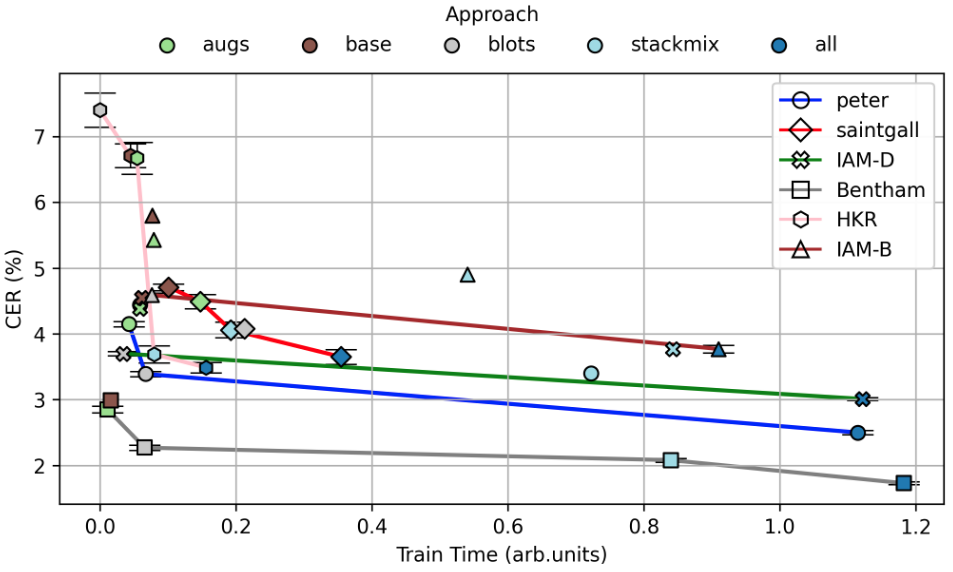}
   \end{tabular}
   \end{center}
   \caption {\label{fig:test_cer}The relative train time and CER results of experiments for different datasets and approaches.}
   \end{figure}

Proposed augmentations allows for metric improvements. In our experiments, we used StackMix ”on the fly” during training. We added traditional augmentations of CLAHE \cite{Reza2004}, JpegCompression, Rotate, and our augmentation (simulation of crossed-out letters) - ”HandWritten Blots”. Different combinations of augmentations were grouped in our experiments (Table \ref{tab:extended-all-results-table}, \ref{tab:advanced-experiments-table}):
\begin{itemize}
    \item "Base" - experiments without augmentations, 300 epoch (HKR 100 epoch)
    \item "Augs" - standart augmentations (CLAHE {\em\cite{Reza2004}}, JpegCompression, Rotate), 300 epoch (HKR 100 epoch)
    \item "Blots" - using only our HandWrittenBlot augmentation, 500 epoch (HKR 150 epoch)
    \item "Stackmix" - using only our Stackmix approach, 1~000 epoch (HKR 300 epoch)
    \item "All" - using all previous augmentations (augs + blots + stackmix), 1~000 epoch (HKR 300 epoch)
\end{itemize}
Models with StackMix were trained during 1~000 epoch, but were not overfitted. We believe they should be trained more with bigger external text corpora. A comparison of our results for various datasets (IAM \cite{marti2002iam}, BenthamR0 \cite{bentham}, Digital Peter \cite{potanin2021digital}, HKR\_Dataset  \cite{nurseitov2020hkr}, Saint Gall \cite{fischer2011transcription}) is presented in Table \ref{tab:extended-all-results-table}, \ref{tab:advanced-experiments-table} and in Figure \ref{fig:test_cer}.

The effect of the proposed two augmentations on quality metrics is shown in Figure \ref{fig:test_cer}. Train time $T_{arb}$ were measured in arbitrary units and obtained by the formula:
\begin{equation}
\small\label{eq7}
T_{arb} = log_2(T/T_{min})
\end{equation} 

Where $T_{min} = 33.6$ ms is the minimum value of train time per one image. The colored lines represent obtained experiment points. Since the quality of approaches grows with increasing train time, some points have no line.

Moreover, we consider 5 more datasets named Konzilsprotokolle C (Konzil), Schiller, Ricordi, Patzig and Schwerin. The datasets were written in Italian modern, and medieval German and were first introduced in the ICFHR 2018 Competition over READ dataset to compare the performance of approaches learning with few labeled pages (as all of these datasets are relatively small). 

In Table \ref{tab:advanced-experiments-table} we provide experimental results for datasets considered above (Konzil, Schiller, Ricordi, Patzig, Schwerin). Despite the small number of samples in each dataset, we succeeded in achieving good metrics. StackMix (and StacMix+Blots+Augmentations which is named "all") leads to significant improvement in recognition quality. This is the main result. StackMix approach helps to generate new training samples, which is vital for few samples datasets. These results show that our model could be applied not only for widespread big datasets but also for less-known small ones. It is enough about twelve translated pages on historical manuscripts to train a good performing model. The "All" approach achieves the best metric value for each dataset. It shows that our augmentations applied to small datasets lead to prosperous and stable training.

\begin{table}[!h]
\caption{\label{tab:extended-all-results-table} Extended results for all experiments with CER/WER/ACC for valid and test partitions.}
\begin{center}

\begin{tabular}{|c|c|c|c|c|c|c|c|c|c|c|c|c|}
\hline
\multirow{2}{*}{} & \multicolumn{6}{c|}{BenthamR0} \\
\cline{2-7}
& Valid CER, \% & Valid WER, \% & Valid ACC, \% & Test CER, \% & Test WER, \% & Test ACC, \% \\
\hline
base & 5.28 ± 0.08 & 26.1 ± 0.1 & 24.9 ± 0.3 & 2.99 ± 0.06 & 11.8 ± 0.3 & 52.1 ± 0.8 \\
augs & 5.28 ± 0.03 & 25.9 ± 0.1 & 25.1 ± 0.4 & 2.85 ± 0.05 & 11.3 ± 0.1 & 52.8 ± 1.0 \\
blots & 4.61 ± 0.02 & 24.4 ± 0.1 & 26.9 ± 0.3 & 2.27 ± 0.04 & 9.5 ± 0.2 & 57.0 ± 0.4 \\
stackmix & 4.51 ± 0.01 & 24.2 ± 0.1 & 26.9 ± 0.4 & 2.08 ± 0.03 & 9.0 ± 0.1 & 58.2 ± 0.8 \\
all & \textbf{4.20 ± 0.03} & \textbf{23.4 ± 0.1} & \textbf{28.1 ± 0.5} & \textbf{1.73 ± 0.02} & \textbf{7.8 ± 0.1} & \textbf{61.9 ± 1.1} \\
\hline
\end{tabular}

\begin{tabular}{|c|c|c|c|c|c|c|c|c|c|c|c|c|}
\hline
\multirow{2}{*}{} & \multicolumn{6}{c|}{IAM-B} \\
\cline{2-7}
& Valid CER, \% & Valid WER, \% & Valid ACC, \% & Test CER, \% & Test WER, \% & Test ACC, \% \\
\hline
base & 3.67 ± 0.07 & 13.2 ± 0.2 & 38.3 ± 0.2 & 5.80 ± 0.08 & 18.9 ± 0.2 & 29.3 ± 0.4 \\
augs & 3.50 ± 0.03 & 12.7 ± 0.2 & 38.8 ± 0.9 & 5.43 ± 0.04 & 17.7 ± 0.1 & 30.7 ± 0.5 \\
blots & 2.93 ± 0.07 & 10.7 ± 0.3 & 44.0 ± 0.7 & 4.59 ± 0.03 & 15.0 ± 0.1 & 36.4 ± 0.5 \\
stackmix & 3.13 ± 0.03 & 11.5 ± 0.0 & 45.4 ± 0.5 & 4.90 ± 0.07 & 16.4 ± 0.2 & 35.2 ± 0.5 \\
all & \textbf{2.40 ± 0.05} & \textbf{8.9 ± 0.2} & \textbf{53.7 ± 1.1} & \textbf{3.77 ± 0.06} & \textbf{12.8 ± 0.2} & \textbf{43.6 ± 0.6} \\
\hline
\end{tabular}

\begin{tabular}{|c|c|c|c|c|c|c|c|c|c|c|c|c|}
\hline
\multirow{2}{*}{} & \multicolumn{6}{c|}{Digital Peter} \\
\cline{2-7}
& Valid CER, \% & Valid WER, \% & Valid ACC, \% & Test CER, \% & Test WER, \% & Test ACC, \% \\
\hline
base & 4.27 ± 0.06 & 22.5 ± 0.3 & 47.1 ± 0.3 & 4.44 ± 0.02 & 24.3 ± 0.2 & 43.7 ± 0.4 \\
augs & 3.89 ± 0.04 & 21.2 ± 0.2 & 49.2 ± 0.5 & 4.15 ± 0.04 & 23.0 ± 0.2 & 45.7 ± 0.4 \\
blots & 3.17 ± 0.06 & 17.3 ± 0.2 & 56.3 ± 0.7 & 3.39 ± 0.04 & 19.3 ± 0.4 & 51.9 ± 0.4 \\
stackmix & 3.19 ± 0.05 & 17.2 ± 0.2 & 55.5 ± 0.4 & 3.40 ± 0.05 & 19.2 ± 0.3 & 51.6 ± 0.7 \\
all & \textbf{2.42 ± 0.08} & \textbf{13.3 ± 0.4} & \textbf{64.0 ± 0.8} & \textbf{2.50 ± 0.03} & \textbf{14.6 ± 0.2} & \textbf{60.8 ± 0.8} \\
\hline
\end{tabular}

\begin{tabular}{|c|c|c|c|c|c|c|c|c|c|c|c|c|}

\hline
\multirow{2}{*}{}& \multicolumn{6}{c|}{IAM-D} \\
\cline{2-7}
& Valid CER, \% & Valid WER, \% & Valid ACC, \% & Test CER, \% & Test WER, \% & Test ACC, \% \\
\hline
base & 3.72 ± 0.03 & 13.5 ± 0.1 & 40.3 ± 0.6 & 4.55 ± 0.06 & 14.5 ± 0.2 & 35.5 ± 0.7 \\
augs & 3.54 ± 0.03 & 12.9 ± 0.2 & 41.9 ± 0.6 & 4.38 ± 0.04 & 14.0 ± 0.2 & 36.3 ± 0.6 \\
blots & 2.94 ± 0.09 & 10.8 ± 0.3 & 46.8 ± 0.9 & 3.70 ± 0.03 & 11.8 ± 0.1 & 42.3 ± 0.6 \\
stackmix & 2.98 ± 0.04 & 11.1 ± 0.1 & 48.0 ± 0.3 & 3.77 ± 0.04 & 12.3 ± 0.1 & 42.4 ± 0.8 \\
all & \textbf{2.32 ± 0.04} & \textbf{8.7 ± 0.1} & \textbf{55.1 ± 0.8} & \textbf{3.01 ± 0.02} & \textbf{9.8 ± 0.1} & \textbf{50.7 ± 0.2} \\
\hline
\end{tabular}

\begin{tabular}{|c|c|c|c|c|c|c|c|c|c|c|c|c|}
\hline
\multirow{2}{*}{} & \multicolumn{6}{c|}{HKR} \\
\cline{2-7}
& Valid CER, \% & Valid WER, \% & Valid ACC, \% & Test CER, \% & Test WER, \% & Test ACC, \% \\
\hline
base & 1.72 ± 0.03 & 5.8 ± 0.1 & 91.1 ± 0.1 & 6.71 ± 0.18 & 22.5 ± 0.3 & 71.1 ± 0.5 \\
augs & 1.78 ± 0.03 & 5.8 ± 0.1 & 91.2 ± 0.1 & 6.67 ± 0.24 & 21.5 ± 0.3 & 72.2 ± 0.1 \\
blots & 1.34 ± 0.04 & 4.1 ± 0.2 & 93.7 ± 0.3 & 7.40 ± 0.26 & 23.0 ± 0.5 & 72.6 ± 0.4 \\
stackmix & 1.72 ± 0.05 & 6.1 ± 0.1 & 90.7 ± 0.1 & 3.69 ± 0.13 & 14.4 ± 0.4 & 80.0 ± 0.3 \\
all & \textbf{1.32 ± 0.02} & \textbf{4.3 ± 0.1} & \textbf{93.4 ± 0.1} & \textbf{3.49 ± 0.08} & \textbf{13.0 ± 0.3} & \textbf{82.0 ± 0.5} \\
\hline
\end{tabular}

\begin{tabular}{|c|c|c|c|c|c|c|c|c|c|c|c|c|}
\hline
\multirow{2}{*}{} & \multicolumn{6}{c|}{Saint Gall} \\
\cline{2-7}
& Valid CER, \% & Valid WER, \% & Valid ACC, \% & Test CER, \% & Test WER, \% & Test ACC, \% \\
\hline
base & 4.61 ± 0.08 & 31.8 ± 0.6 & 3.0 ± 0.6 & 4.71 ± 0.05 & 32.5 ± 0.3 & 2.2 ± 0.5 \\
augs & 4.39 ± 0.09 & 30.7 ± 0.5 & 3.8 ± 0.5 & 4.49 ± 0.11 & 31.3 ± 0.5 & 3.4 ± 0.7 \\
blots & 4.12 ± 0.04 & 28.6 ± 0.3 & 5.4 ± 1.1 & 4.08 ± 0.03 & 28.1 ± 0.2 & 4.6 ± 1.1 \\
stackmix & 4.11 ± 0.06 & 29.1 ± 0.4 & 7.5 ± 0.9 & 4.06 ± 0.12 & 28.8 ± 0.8 & 6.1 ± 0.8 \\
all & \textbf{3.73 ± 0.06} & \textbf{26.5 ± 0.3} & \textbf{12.0 ± 2.5} & \textbf{3.65 ± 0.11} & \textbf{26.2 ± 0.6} & \textbf{11.8 ± 2.0} \\
\hline
\end{tabular}

\begin{tabular}{|c|c|c|c|c|c|c|c|c|c|c|c|c|}
\hline
\multirow{2}{*}{} & \multicolumn{6}{c|}{Washington CV1} \\
\cline{2-7}
& Valid CER, \% & Valid WER, \% & Valid ACC, \% & Test CER, \% & Test WER, \% & Test ACC, \% \\
\hline
base & 6.20 ± 0.40 & 24.4 ± 1.4 & 19.2 ± 3.0 & 7.05 ± 0.58 & 25.1 ± 1.5 & 18.6 ± 2.8 \\
augs & 6.35 ± 0.68 & 24.3 ± 2.1 & 19.2 ± 3.3 & 6.64 ± 0.56 & 24.0 ± 1.2 & 19.1 ± 5.1 \\
blots & 3.87 ± 0.23 & 16.0 ± 0.8 & 32.6 ± 3.2 & 4.03 ± 0.16 & 14.6 ± 0.8 & 38.4 ± 3.1 \\
stackmix & 3.65 ± 0.07 & 16.4 ± 0.2 & 33.3 ± 1.9 & 5.23 ± 0.08 & 20.8 ± 0.7 & 28.6 ± 1.2 \\
all & \textbf{2.54 ± 0.05} & \textbf{11.8 ± 0.2} & \textbf{42.5 ± 1.2} & \textbf{3.66 ± 0.10} & \textbf{14.4 ± 0.6} & \textbf{41.1 ± 2.1} \\
\hline
\end{tabular}

\end{center}

\end{table}

\begin{table}[!h]
\caption{\label{tab:advanced-experiments-table} Extended results for ICFHR 2018 Competition over READ dataset with CER/WER/ACC for valid and test partitions.}
\begin{center}

\begin{tabular}{|c|c|c|c|c|c|c|c|c|c|c|c|c|}
\hline
\multirow{2}{*}{}& \multicolumn{6}{c|}{Konzil} \\
\cline{2-7}
& Valid CER, \% & Valid WER, \% & Valid ACC, \% & Test CER, \% & Test WER, \% & Test ACC, \% \\
\hline
base & 9.40 ± 0.34 & 40.3 ± 2.0 & 26.9 ± 2.9 & 10.46 ± 0.48 & 42.6 ± 1.1 & 21.8 ± 2.9 \\
augs & 8.39 ± 1.29 & 36.4 ± 3.9 & 30.3 ± 2.9 & 10.41 ± 1.49 & 41.1 ± 3.0 & 24.4 ± 2.5 \\
blots & 5.86 ± 0.47 & 27.6 ± 1.5 & 36.5 ± 3.1 & 7.38 ± 0.70 & 31.8 ± 2.2 & 30.3 ± 2.5 \\
stackmix & 4.20 ± 0.39 & 23.9 ± 2.6 & 41.4 ± 4.2 & 5.60 ± 0.20 & 28.7 ± 0.4 & 32.4 ± 1.5 \\
all & \textbf{2.85 ± 0.24} & \textbf{16.9 ± 1.0} & \textbf{48.3 ± 2.4} & \textbf{3.31 ± 0.24} & \textbf{17.4 ± 1.4} & \textbf{46.7 ± 4.1} \\
\hline
\end{tabular}

\begin{tabular}{|c|c|c|c|c|c|c|c|c|c|c|c|c|}
\hline
\multirow{2}{*}{}& \multicolumn{6}{c|}{Ricordi} \\
\cline{2-7}
& Valid CER, \% & Valid WER, \% & Valid ACC, \% & Test CER, \% & Test WER, \% & Test ACC, \% \\
\hline
base & 49.39 ± 12.46 & 86.3 ± 9.6 & 2.1 ± 2.9 & 50.79 ± 12.07 & 88.6 ± 7.7 & 0.0 ± 0.0 \\
augs & 18.88 ± 3.28 & 56.9 ± 6.1 & 2.1 ± 2.9 & 19.84 ± 2.04 & 58.1 ± 3.3 & 0.0 ± 0.0 \\
blots & 49.77 ± 16.67 & 84.3 ± 12.1 & 2.1 ± 2.9 & 52.59 ± 15.64 & 88.2 ± 9.8 & 0.0 ± 0.0 \\
stackmix & 10.73 ± 1.55 & 39.0 ± 4.0 & 6.3 ± 2.4 & 12.25 ± 0.90 & 40.6 ± 2.1 & 2.3 ± 2.2 \\
all & \textbf{9.89 ± 2.75} & \textbf{36.6 ± 6.7} & \textbf{9.2 ± 5.0} & \textbf{11.54 ± 2.00} & \textbf{38.4 ± 5.7} & \textbf{3.3 ± 3.2} \\
\hline
\end{tabular}

\begin{tabular}{|c|c|c|c|c|c|c|c|c|c|c|c|c|}
\hline
\multirow{2}{*}{}& \multicolumn{6}{c|}{Schiller} \\
\cline{2-7}
& Valid CER, \% & Valid WER, \% & Valid ACC, \% & Test CER, \% & Test WER, \% & Test ACC, \% \\
\hline
base & 14.60 ± 2.04 & 45.7 ± 3.9 & 2.9 ± 2.6 & 17.31 ± 1.17 & 58.5 ± 2.5 & 0.0 ± 0.0 \\
augs & 14.51 ± 3.20 & 44.9 ± 5.7 & 3.8 ± 4.0 & 16.35 ± 1.83 & 54.7 ± 2.9 & 0.0 ± 0.0 \\
blots & 8.98 ± 0.68 & 32.5 ± 2.5 & 13.3 ± 4.0 & 10.98 ± 0.64 & 41.3 ± 1.5 & 1.9 ± 1.3 \\
stackmix & 6.15 ± 0.29 & 27.2 ± 1.3 & 19.1 ± 5.8 & 8.42 ± 0.23 & 35.5 ± 1.3 & 5.4 ± 2.1 \\
all & \textbf{3.68 ± 0.39} & \textbf{17.5 ± 1.2} & \textbf{27.6 ± 2.1} & \textbf{5.79 ± 0.31} & \textbf{26.1 ± 0.8} & \textbf{16.2 ± 2.1} \\
\hline
\end{tabular}

\begin{tabular}{|c|c|c|c|c|c|c|c|c|c|c|c|c|}
\hline
\multirow{2}{*}{}& \multicolumn{6}{c|}{Schwerin} \\
\cline{2-7}
& Valid CER, \% & Valid WER, \% & Valid ACC, \% & Test CER, \% & Test WER, \% & Test ACC, \% \\
\hline
base & 7.32 ± 0.00 & 27.2 ± 0.0 & 17.6 ± 0.0 & 8.65 ± 0.00 & 30.1 ± 0.0 & 14.3 ± 0.0 \\
augs & 16.56 ± 4.87 & 46.4 ± 12.0 & 3.2 ± 7.2 & 17.98 ± 4.47 & 49.5 ± 9.6 & 5.1 ± 4.6 \\
blots & 2.26 ± 0.00 & 9.1 ± 0.0 & 63.2 ± 0.0 & 3.28 ± 0.00 & 12.6 ± 0.0 & 53.6 ± 0.0 \\
stackmix & 2.13 ± 0.20 & 8.6 ± 0.6 & 61.2 ± 2.2 & 3.09 ± 0.25 & 12.8 ± 1.3 & 52.9 ± 3.4 \\
all & \textbf{1.92 ± 0.11} & \textbf{7.6 ± 0.3} & \textbf{65.0 ± 1.9} & \textbf{2.91 ± 0.08} & \textbf{12.2 ± 0.3} & \textbf{53.5 ± 0.9} \\
\hline
\end{tabular}

\begin{tabular}{|c|c|c|c|c|c|c|c|c|c|c|c|c|}
\hline
\multirow{2}{*}{}& \multicolumn{6}{c|}{Patzig} \\
\cline{2-7}
& Valid CER, \% & Valid WER, \% & Valid ACC, \% & Test CER, \% & Test WER, \% & Test ACC, \% \\
\hline
base & 37.15 ± 1.90 & 77.0 ± 1.9 & 2.6 ± 0.0 & 37.92 ± 1.74 & 79.8 ± 1.4 & 0.2 ± 0.4 \\
augs & 34.76 ± 4.84 & 75.0 ± 5.4 & 2.1 ± 1.2 & 35.32 ± 5.02 & 76.3 ± 3.9 & 0.5 ± 0.8 \\
blots & 31.25 ± 1.73 & 71.5 ± 2.1 & 3.2 ± 1.2 & 32.70 ± 1.81 & 73.0 ± 1.8 & 0.0 ± 0.0 \\
stackmix & 14.97 ± 0.86 & 42.7 ± 1.8 & 11.6 ± 1.4 & 13.72 ± 0.61 & 46.5 ± 1.8 & 7.3 ± 2.3 \\
all & \textbf{12.80 ± 1.25} & \textbf{37.3 ± 3.0} & \textbf{14.5 ± 3.4} & \textbf{11.34 ± 0.39} & \textbf{38.6 ± 0.3} & \textbf{13.6 ± 1.0} \\
\hline
\end{tabular}

\end{center}

\end{table}

\subsection{Ablation study}
\label{sec:sup}

In this section, we present the comparison with other models (Table \ref{tab:hwb-result-comparison}).Our model outperforms other approaches on the BenthamR0, HKR, and IAM-D datasets. It reached 3.77\% of CER on the IAM-B dataset, which is very close to the best model published in 2016 that achieved 3.5\% CER. On the Saint Gall dataset, we achieved 5.56\% CER, which is very close to the current best solution of 5.26\% CER.

IAM dataset has two versions because  papers use various data splits. We included IAM-B and IAM-D partitions in Table \ref{tab:hwb-result-comparison} to compare them with other models of the same split.

Authors of the paper \cite{de2020htr} provided open code for their model here \cite{htrflorlink}. We noticed that when evaluating the model, they lowered the characters in predicted and true strings. However, in our experiments, we did not convert the characters to lowercase. For real HTR tasks, it is not important to track the case of characters. 

We compare our approach on a few labeled datasets (Konzil, Schiller, Ricordi, Patzig, Schwerin) with results from the paper \cite{aradillas2020boosting}. Results are presented in Table \ref{tab:read-comparison}. For our model, we took the best results from Table \ref{tab:advanced-experiments-table} and also we took the best results from Table 1 in paper \cite{aradillas2020boosting}. Our model achieves better CER for 3 of 5 datasets. Additionally, we did not use transfer learning from bigger datasets, unlike authors of \cite{aradillas2020boosting}.

Data which we used for Table \ref{tab:read-comparison} are located in "specific\_data" folder. Each image and text translation has index "train\_[1,4,16]" in their file names. So we set "train\_1" as a valid index, "train\_4" as test index, and "train\_16" as train index. We did it this way because "train\_1" has the smallest number of samples for each dataset, "train\_4" is slightly bigger, and train\_16 is the largest. So for the test set in Table \ref{tab:read-comparison}, we took all files marked with "train\_4" for each dataset. But authors of \cite{aradillas2020boosting} used the test set (with ground truth) for ICFHR 2018 Competition \cite{strauss2018icfhr2018}, which is hidden for us. Despite the usage of different test sets, we think this comparison is relatively fair.

\begin{table}[!h]
\caption{\label{tab:hwb-result-comparison} Comparison to other models, test set (IAM, Bentham, HKR, Saint Gall, Digital Peter).}
\begin{center}
\begin{tabular}{|c|c|c|}
\hline
\multirow{2}{*}{Model} & \multicolumn{2}{c|}{IAM-B}\\
\cline{2-3}
& CER, \% & WER, \% \\
\hline
\cite{voigtlaender2016handwriting} & \textbf{3.5} & \textbf{9.3} \\
\cite{yousef2020origaminet} & 4.7 & - \\
\cite{coquenet2020end} & 4.32 & 16.24 \\
\cite{coquenet2020recurrence} & 7.99 & 28.61\\
\cite{moysset20192d} & 7.73 & 25.22\\
\cite{wang2020decoupled} & 6.64 & 19.6\\
ours & 3.77  & 12.8\\
\hline
\multirow{2}{*}{} & \multicolumn{2}{c|}{IAM-D}\\
\cline{2-3}
\cline{2-3}
\cite{abdallah2020attention} &7.8 & 25.5\\
ours & \textbf{3.01}  & \textbf{9.8}\\

\hline
\multirow{2}{*}{}& \multicolumn{2}{c|}{Digital Peter}\\
\cline{2-3}
\cite{complink} & 10.5 & 44.4\\
ours & \textbf{2.50}  & \textbf{14.6}\\

\hline
\multirow{2}{*}{} & \multicolumn{2}{c|}{BenthamR0}\\
\cline{2-3}
\cite{abdallah2020attention} & 7.1 & 20.9 \\
\cite{de2020htr} & 3.98  & 9.8\\
ours & \textbf{1.73}  & \textbf{7.9}\\

\hline
\multirow{2}{*}{} & \multicolumn{2}{c|}{HKR}\\
\cline{2-3}
\cite{abdallah2020attention} & 4.5 & 19.2\\
ours & \textbf{3.49}  & \textbf{13.0}\\

\hline
\multirow{2}{*}{}& \multicolumn{2}{c|}{Saint Gall}\\
\cline{2-3}
\cite{abdallah2020attention} & 7.25 & 23.0\\
\cite{de2020htr}  & 5.26 & \textbf{21.14} \\
ours & \textbf{3.65}  & 26.2 \\
\hline
\end{tabular}
\end{center}
\end{table}

\begin{table}[!h]
\caption{Comparison to other models, test CER for ICFHR 2018 Competition over READ dataset.} 
\label{tab:read-comparison}
\begin{center}
\begin{tabular}{|c|c|c|}
\hline
\multirow{3}{*}{Dataset} & \multicolumn{2}{c|}{Model}\\
\cline{2-3}
& CNN-BLSTM & \\
&(bidirectional LSTM)
& ours \\
\hline
Konzil& 4.37 [4.24-4.54] & \textbf{3.31 ± 0.24} \\
Ricordi & \textbf{11.2 [10.11-11.23]} & 11.54 ± 2.00\\
Schiller & 9.4 [9.31-9.46] & \textbf{5.79 ± 0.31} \\
Schwerin & 3.5 [3.46-3.53] & \textbf{2.91 ± 0.08}\\
Patzig & \textbf{10.6 [10.52-10.65]} & 11.34 ± 0.39 \\
\hline
\end{tabular}
\end{center}
\end{table}

\section{CONCLUSION}
We have introduced two new data augmentation techniques, strikethrough text algorithm HandWritten Blots and handwritten text generation algorithm StackMix, based on weakly-supervised training. We have demonstrated their use with Resnet - BiLSTM - CTC network to produce the best result among the currently known handwriting recognition systems. These techniques produce the lowest word error rate (WER) and Character Error Rate (CER) to date over hundreds of authors, multiple languages, and thousands of documents, including challenging, medieval, historical documents with noise, ink bleed-through, and faint handwriting.

The presented system can significantly increase the speed of deciphering historical documents. For example, it took a team of 10-15 historians about three months to decipher 662 pages of manuscripts from the Digital Peter dataset. When working on the same dataset on a single Tesla V100, the average decryption speed was 95 lines/s or 380 pages/min, unattainable by historical scientists.



\bibliographystyle{unsrt}
\bibliography{icmv} 

\end{document}